\documentclass[twocolumn, 12, conference]{article}
\usepackage{times}  
\usepackage{helvet}  
\usepackage{courier}  
\usepackage[hyphens]{url}  
\usepackage{graphicx} 
\usepackage{amsmath}
\usepackage{amsfonts} 
\usepackage{verbatim}
\usepackage{hyperref}
\usepackage{natbib}  
\usepackage{caption} 
\usepackage{float}
\usepackage{algorithm}
\usepackage{algorithmic}
\usepackage{authblk}
\usepackage{newfloat}
\usepackage{listings}
\DeclareCaptionStyle{ruled}{labelfont=normalfont,labelsep=colon,strut=off} 
\floatstyle{ruled}
\newfloat{listing}{tb}{lst}{}
\floatname{listing}{Listing}

\title{LaGR-SEQ: Language-Guided Reinforcement Learning with Sample-Efficient Querying}
\author{Thommen George Karimpanal, Laknath Buddhika Semage, Santu Rana,\\ Hung Le, Truyen Tran, Sunil Gupta, Svetha Venkatesh}
\affil{Applied Artificial Intelligence Institute, Deakin University, Australia}

\begin{document}

\maketitle

\begin{abstract}
Large language models (LLMs) have recently demonstrated their impressive ability to provide context-aware responses via text. This ability could potentially be used to predict plausible solutions in sequential decision making tasks pertaining to pattern completion.
For example, by observing a partial stack of cubes, LLMs can predict the correct sequence in which the remaining cubes should be stacked by extrapolating the observed patterns (e.g., cube sizes, colors or other attributes) in the partial stack. 
In this work, we introduce LaGR (\emph{Language-Guided Reinforcement learning}), which uses this predictive ability of LLMs to propose solutions to tasks that have been partially completed by a primary reinforcement learning (RL) agent, in order to subsequently guide the latter's training. However, as RL training is generally not sample-efficient, deploying this approach would inherently imply that the LLM be repeatedly queried for solutions; a process that can be expensive and infeasible. To address this issue, we introduce SEQ (\emph{sample efficient querying}), where we simultaneously train a secondary RL agent to decide when the LLM should be queried for solutions. Specifically, we use the quality of the solutions emanating from the LLM as the reward to train this agent. We show that our proposed framework LaGR-SEQ enables more efficient primary RL training, while simultaneously minimizing the number of queries to the LLM. We demonstrate our approach on a series of tasks and highlight the advantages of our approach, along with its limitations and potential future research directions.
\end{abstract}

\section{Introduction}

Reinforcement Learning (RL) \cite{sutton1998introduction} agents have demonstrated their ability to learn a variety of tasks ranging from video game playing \cite{mnih2015human,mnih2013playing} to robotics \cite{singh2022reinforcement,kober2013reinforcement}, through the grounded approach of interacting with the environment, with the aim of maximizing their long term cumulative rewards. However, the underlying learning mechanism fundamentally involves trial and error of different actions without context, which is expensive and wasteful, particularly in real world scenarios. 
To address this limitation, we aim to leverage recently proposed Large Language Models (LLMs) such as GPT-3 \cite{brown2020language} which are capable of extracting contextually meaningful information from a wide variety of scenarios. 

\begin{figure}[h]
    \includegraphics[width=\columnwidth]{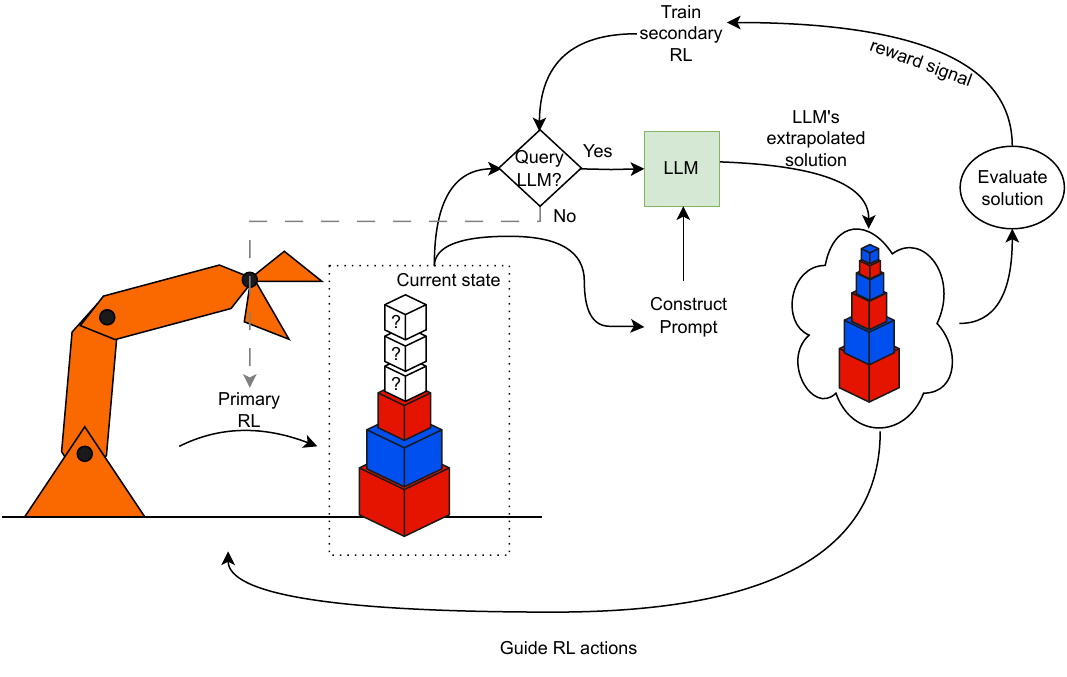}
    \caption{The overall flow of our approach. A primary RL agent interacts with the environment to solve a task. The LLM, prompted with the current state, produces an extrapolated solution which guides the exploration of the primary RL agent. The evaluation of the LLM solution is used as a reward signal to simultaneously train a secondary RL agent, which determines whether the LLM should be queried.}
    \label{llmflow}
\end{figure}

LLMs have previously been used to provide RL agents with contextual information. In \citet{du2023guiding}, RL agents were rewarded for achieving goals suggested by LLMs when prompted with the agent's state. Such agents were shown to exhibit improved performance in a variety of downstream tasks owing to their context-informed pretraining. In a similar spirit, we use state-based prompting in pattern completion RL tasks, to use LLMs to predict plausible complete patterns by prompting it with intermediate states that occur during learning, with the intention of using this prediction to accelerate RL training. We also show how RL can in turn be used to query LLMs efficiently, thus laying the foundation for future works aiming to efficiently integrate LLMs' context richness into RL tasks. 


Specifically, we propose two mechanisms: (1) LaGR, in which a \emph{primary} RL agent that learns to solve the task through environment interactions, while being guided by the extrapolated solutions from an LLM, and (2) SEQ (\emph{Sample Efficient Querying}), in which a \emph{secondary} RL agent learns to query the LLM only in those scenarios where its solutions are likely to be correct, thereby minimizing the number of LLM queries. Together, our LaGR-SEQ framework aims to simultaneously achieve improved sample efficiency in learning the task policy, as well as efficient querying of the LLM.

To illustrate our idea, we refer to Figure \ref{llmflow}, where we depict a stacking robot as a \emph{primary} RL agent, which aims to stack six boxes in a specific target pattern. The challenge for RL agents in such tasks is that they lack contextual information (e.g. the concept of alternating box colors or decreasing or increasing box sizes), and as such, they must initially explore actions randomly, in an undirected and uninformed manner, learning to assign higher values to actions that lead to more desirable stack configurations, until the boxes are eventually stacked in the desired pattern. In contrast to this, we posit that LLMs are capable of leveraging contextual knowledge to extract patterns in the intermediate states during RL training (i.e., patterns such as alternating box colors and decreasing box sizes in Figure \ref{llmflow}) and extrapolate them to produce plausible solutions to the task.
 We in turn use these extrapolated solutions from the LLM to guide the \emph{primary} RL agent's actions, thereby leading to significant improvements in its sample efficiency. We refer to this mechanism as \emph{Language-Guided Reinforcement learning} or LaGR. We note that LLMs alone would not fare well on these tasks, as without the in-context prompting (partial solutions) facilitated by the RL agent, the space of possible patterns would be large, thus making LLM an ineffective predictor. On the other hand, although RL alone could possibly solve this task, the approach would be significantly less efficient compared to our proposed LaGR mechanism.

Despite the potential benefits of accelerated RL training through the described LaGR mechanism, it inherently involves repeatedly querying the LLM for suggested solutions. As RL training typically involves several thousands of interactions, it is infeasible and expensive to query the LLM throughout training. Instead, we evaluate the proposed LLM solution, and use it as a reward signal for a \emph{secondary} RL agent, which is trained to decide whether or not an LLM query should be made in a given state. This agent eventually learns to query the LLM only in states where the partial solution elicits correct responses from the LLM, thereby obviating LLM queries in all other scenarios. We refer to this mechanism as \emph{Sample Efficient Querying} or SEQ. Notably, both mechanisms operate simultaneously (LaGR-SEQ), in parallel, without the need for dedicated interactions for each.

To summarize, the main contributions of this work are:

\begin{itemize}
    \item An integrated framework LaGR-SEQ, to leverage the context-awareness of LLMs for RL training, while simultaneously minimizing the number of LLM queries.
    \item LaGR, a simple approach to guide RL actions in pattern completion tasks using extrapolated solutions from the LLM, leading to faster learning of the desired policy.
    \item SEQ, an automated mechanism to determine when the LLM should be queried.
    \item Empirical evaluations of our proposed framework, which demonstrate its ability to excel in pattern/sequence completion tasks where neither RL nor LLMs perform well.
\end{itemize}

\section{Related Work}
Recent LLMs \cite{brown2020language} developed with the aid of human feedback \cite{christiano2017deep} have demonstrated their ability to produce context-aware and meaningful responses in several domains \cite{mirchandani2023large}, which has propelled their widespread adoption for a number of applications \cite{yuan2022wordcraft,meyer2023chatgpt,kasneci2023chatgpt}. These context-aware capabilities of LLMs were also found to be useful for RL agents, which were typically set up to solve sequential decision making problems devoid of context. In order to further strengthen context-awareness, SayCan \cite{ahn2022can} demonstrated effective language grounding using pretrained behaviors, while PaLM-E \cite{driess2023palm} showed multimodal embodied learning by feeding a robot's sensory data into the language model. Other approaches \cite{huang2022language,li2022pre} augmented RL agents with LLMs, primarily to ground high level tasks to low level actions, with the use of demonstrations. Apart from these, a number of works in the past \cite{shridhar2020alfred,lynch2020language,anderson2018vision} have focused on solving RL tasks by directly conditioning policies on language. 

\citet{lin2023learning} recently proposed Dynalang, a framework to help agents predict the future representations of the environment state, which in turn provides a rich learning signal for agents interacting with the world. \citet{du2023guiding} proposed an approach to pretrain RL agents through an intrinsic reward mechanism, on potential future tasks inferred via LLM so that the RL agent is better prepared for downstream tasks. Similar to \citet{du2023guiding}, \citet{choi2022lmpriors} proposed LMPriors to construct task specific priors with a focus on improving downstream model performance. \citet{kant2022housekeep} also demonstrated the ability of LLMs to use contextual knowledge to help arrange unseen objects in unknown environments in a housekeeping task. Although these works essentially extrapolate future tasks with a focus on downstream tasks, they differ from our work, which focuses on the extrapolation of the final state configuration within a task, prompted by partially complete intermediate states that are encountered during RL training. 

In \citet{kwon2023reward}, LLMs were used to generate binary rewards in a negotiation task based on the style of negotiation. In a similar spirit, our \emph{secondary} RL agent (SEQ) is also trained based on the LLM's responses, although we evaluate the accuracy of the responses with respect to the task at hand, and use those evaluation scores as a reward signal for training SEQ.
Moreover, in contrast to \citet{kwon2023reward}, the rewards used in our work could potentially also be non-binary. We now describe the details of our proposed approach in the subsequent sections.

\section{Framework and Methodology}
Our proposed LaGR-SEQ framework consists of a \emph{primary} RL agent, a pretrained LLM and a \emph{secondary} RL agent, which we describe in detail in this section.\\
\emph{Primary RL Agent:} 
We consider pattern completion as a sequential decision making problem, modeled as a Markov decision process (MDP) \cite{puterman2014markov} $\mathcal{M=\{S,A,R,T\}}$, where $\mathcal{S}$ is the state space, $\mathcal{A}$ is the action space, $\mathcal{R}:\mathcal{S}\times\mathcal{A}\rightarrow \mathbb{R}$ is the reward function and $\mathcal{T}:\mathcal{S}\times\mathcal{A}\rightarrow\mathcal{S}$ is the transition function. We assume each state $s\in \mathcal{S}$ corresponds to a configuration (of say, objects $o \in O$, where $O$ is the set of all objects), and there exists some state $s^*$ corresponding to the target object configuration/pattern. An action $a\in \mathcal{A}$ is any operation that transforms state $s$ to a new state $s'$. We aim to learn a policy $\pi^*$ that learns to produce the target pattern $s^*$ by maximizing the accumulated rewards during the \emph{primary} agent's interaction with the environment. 

\noindent\emph{Pretrained LLM:} 
We assume the availability of a pretrained LLM $\phi=\{\psi,\xi\}$ which extrapolates both a policy $\psi:\mathcal{D}\times \mathcal{S}\rightarrow \Pi$ ($\Pi$ is the space of policies) as well the resulting configuration $\xi:\mathcal{D}\times \mathcal{S}\rightarrow \mathcal{S}$ by taking task descriptor $d\in\mathcal{D}$ ($\mathcal{D}$ is the description space) and the current state/configuration as input. We refer to the resulting extrapolated policy as $\pi_{LLM}$, and the resulting extrapolated state as $s_{LLM}$. That is:

\begin{equation}
    \pi_{LLM}=\psi(d,s)
\end{equation}

\begin{equation}
    s_{LLM}=\xi(d,s)
\end{equation}
The task descriptor $d$ specifies the context for the LLM, and may contain a description of the task at hand, the specific representation of $\pi_{LLM}$ and $s_{LLM}$, with examples of similar tasks along with their corresponding solutions, either with or without accompanying explanations for the solutions. In our setting, we directly infer $\pi_{LLM}$ from $s_{LLM}$. For example, if $s_{LLM}$ is a stack of boxes arranged according to the colors of the rainbow (VIBGYOR), $\pi_{LLM}$ would then correspond to stacking a Violet box, followed by an Indigo one, and so on, with the action being a choice of box color.  The task descriptor $d$ together with the current state $s$ composes the prompt for the LLM. We note that a poorly designed prompt may return undesirable LLM responses, and as such, it may require considerable effort to design an appropriate prompt. In this work, we use a fixed $d$, and focus on improving the LLM response by allowing the \emph{primary} RL agent to feed more appropriate states $s$ to the LLM, that are more likely to elicit the desired response from it.

\noindent\emph{Secondary RL Agent:} 
Our framework introduces the \emph{secondary} RL agent to improve the efficiency of querying the LLM. This agent is tasked with determining whether the LLM should be queried from a given state or not. As such, the querying policy consists of the decision of whether to query the LLM at each state, which is equivalent to solving a multi-armed bandit (MAB) \cite{slivkins2019introduction} problem. As such, the \emph{secondary} agent is essentially a classifier that we train to distinguish states that lead to accurate LLM responses from those that do not. However, in this work, we choose to train this classifier using RL. 

We base the \emph{secondary} RL agent on the MDP $\bar{\mathcal{M}}=\{\mathcal{S}, \bar{\mathcal{A}},\bar{\mathcal{R}},\bar{\mathcal{T}}\}$, which shares the same state space as the primary agent. The action space for this agent is binary, corresponding to binary values of action $\bar{a}\in\bar{\mathcal{A}}$ to represent the decision to query or not query the LLM.  As we consider the MAB setting, the transition function $\bar{\mathcal{T}}$ is such that every action leads to a terminal state.
The reward function $\bar{\mathcal{R}}:\mathcal{S}\times \bar{\mathcal{A}}\times\mathcal{D}\rightarrow\mathbb{R}$ is specified as:

\begin{equation}
  \text{$\bar{r}(s,\bar{a},d$)} =
    \begin{cases}
      \text{$0$} & \text{if $\bar{a}=0$}\\
      \text{$E(\xi(d,s)$}) & \text{if $\bar{a}=1$}
    \end{cases}       
\end{equation}

\noindent where $\bar{r}\in\bar{\mathcal{R}}$ and $E:\mathcal{S}\rightarrow\mathbb{R}$ is a function that evaluates the quality of the LLM solution. This evaluation function may return either binary or continuous values, with higher values for LLM predicted configurations $\xi(d,s)$ that are closer to the goal state. We note that although $\mathcal{R}$ could itself be used as the evaluation function $E$, but in general, these functions may be distinct from each other. 

\noindent\subsection{LaGR: Guiding RL with LLM} Our overall architecture begins with the \emph{primary} RL agent interacting with the environment with the aim of maximizing the expected sum of rewards, as specified by $\mathcal{R}$. The nature of the reward function $\mathcal{R}$ is such that state-action pairs that lead to configurations that are closer to the target pattern receive small positive rewards, while those that lead to configurations further away from it receive low rewards. Naturally, as the \emph{primary} agent interacts with the environment, the configurations get closer to the target configuration/pattern. Once LLM querying is triggered, based on the current state $s$, the LLM produces $\pi_{LLM}$ and $s_{LLM}$ through extrapolation. We note that this extrapolated state $s_{LLM}$ is evaluated as $E(s_{LLM})$, and is considered to correspond to a correct solution only when $E(s_{LLM})>\delta$ (where $\delta$ is a threshold). $E(s_{LLM})\leq\delta$ implies that the information contained within $s$ may not have been sufficient for the LLM to extrapolate accurate solutions. If this is the case, the \emph{primary} RL continues interacting with the environment until the LLM is queried again. In this work, we trigger LLM querying each time the \emph{primary} agent experiences a positive reward. The intuition is that positive rewards may indicate an improvement in configuration, which may make it more likely for LLM to extrapolate the target pattern. As such, we note that the pattern completion task can potentially be completed by the \emph{primary} RL agent alone, although it would require a large number of environment interactions.

The contextual knowledge embedded within the LLM on the other hand, could potentially identify the patterns within $s$ and successfully extrapolate it (i.e., $E(s_{LLM})>\delta$).
In such a case, the $\emph{primary}$ agent simply uses the corresponding LLM policy $\pi_{LLM}=\psi(d,s)$ to guide its actions. Specifically, the RL agent selects with high probability $P$, the actions suggested by $\pi_{LLM}$, while simultaneously updating its action-value function. Due to this guidance from $\pi_{LLM}$, the \emph{primary} agent is trained in a more task-oriented manner, achieving high returns without excessive wasteful environment interactions. While this enables good performance in the task, it also limits the agent's exposure to a large portions of the state-action space which are not associated with the solution. Hence, in order to train a more robust agent, we suggest the LLM be followed with a probability $P<1$, with lower values implying less reliance on the LLM and poorer performance, but better state-action coverage.

\noindent\subsection{SEQ: Improving Querying Efficiency}

The LaGR mechanism described in the previous subsection demands that the LLM be queried each time the primary agent achieves a positive reward. However, we posit that from some states/configurations, the LLM may not be able to extrapolate the target pattern, as these states may contain insufficient contextual information to do so. Querying from such states may unnecessarily incur cost and would be wasteful, as the resulting LLM response would probably be inaccurate. Incorrect responses could also occur due to imperfect prompting or due to high LLM temperatures. In order to ensure the LLM is queried only from states where there is a high probability of producing correct solutions, we employ the \emph{secondary} RL agent, which learns to decide whether to query the LLM from a given state or not. 

We train the \emph{secondary} RL agent based on a reward function that depends on the evaluation of the LLM response $E(\xi(d,s)$, such that the act of querying from a state $s$ would be rewarded with high values if the LLM produces desirable responses from that state, and correspondingly low (including negative) reward values for poorer LLM responses. Hence, the \emph{secondary} RL agent eventually learns to query the LLM only at those states where the action-value is relatively higher for the querying action compared to that of the non-querying action. For non-querying actions, the \emph{secondary} RL agent receives a neutral reward (say, $0$) from the evaluation function $E$. We also design $E$ to produce a high (positive) value when $s_{LLM}$ closely matches the target pattern/configuration and low (negative) values when $s_{LLM}$ is far from ideal. We note that in practice, non-binary forms of reward may be more effective at improving querying efficiency (refer to Appendix), although for simplicity, in most of our experiments, we employ the binary variant. In either case, if the LLM, when queried from state $s$ produces $s_{LLM}$ associated with a low evaluation score, the \emph{secondary} RL agent discourages querying from $s$. The \emph{secondary} RL agent thus facilitates sample efficient querying (SEQ) of the LLM. 

When put together, LaGR-SEQ enables faster RL training by extracting contextual information from the LLMs with a minimal number of queries to the LLM. The overall approach is summarized in Algorithm \ref{alg:algo}.

\begin{algorithm}[h]
\caption{LaGR-SEQ: Overall Algorithm}\label{alg:algo}
\begin{algorithmic}

\STATE {Initialize $Q-$ functions $Q_{primary}$ and $Q_{secondary}$, LLM $\phi=\{\psi, \xi\}$, Evaluation function $E$, initial state $s$, maximum episodes $N$, episode horizon $H$, Solution threshold $\delta$, Solution flag $F=0$, \emph{primary} reward initialized  $r=0$}
\FOR{$n=0:N$}
\FOR{$h=0:H$}
\IF{$F=0$}
\STATE{Choose $\bar{a}$ (e.g., $\bar{a}\sim\epsilon-$greedy w.r.t. $Q_{secondary}$)}
\IF{$r>0$ and $\bar{a}=1$}

\STATE{Query LLM: $\pi_{LLM}=\psi(d,s)$; $s_{LLM}=\xi(d,s)$}

\STATE{Evaluate LLM solution as: $\bar{r}=E(s_{LLM})$}

\IF{$E(s_{LLM})>\delta$}
\STATE{set $F=1$}
\ENDIF
\ENDIF
\STATE {Select primary action $a$ (e.g., $a\sim\epsilon-$greedy w.r.t. $Q_{primary}$)}
\STATE {$\bar{r}=0$}
\ELSE
\STATE {With high probability $P$,  $a\sim\pi_{LLM}$ (sample \emph{primary} RL actions from $\pi_{LLM}$)}
\ENDIF
\STATE {Take action $a$, get $r$ and $s'$}
\STATE {Update $Q_{primary}$ using ($s,a,s',r$)}
\STATE {Update $Q_{secondary}$ using ($s,\bar{a},\bar{r}$)}
\STATE $s\gets s'$
\ENDFOR
\ENDFOR
\end{algorithmic}
\end{algorithm}

\section{Experiments}
We empirically evaluate several aspects of our proposed LaGR-SEQ framework. Specifically we aim to evaluate: (a) Can LLMs extrapolate patterns? (b) Does LaGR-SEQ help improve the sample efficiency of RL training? (c) Does SEQ help reduce the number of LLM queries? (d) How does the selected LLM variant and LLM temperature affect LLM prediction accuracy and LaGR/SEQ? 

\subsection{Environments:}
We consider cube-stacking, image completion and object arrangement tasks to empirically evaluate our approach\footnote{code and data can be found \href{https://github.com/GKthom/LaGRSEQ}{here} }. We use custom environments due to the unavailability of standard benchmarks for RL based pattern completion tasks relevant to the context of this work. We describe these environments in detail and provide in the Appendix the task descriptor $d$ along with other hyperparameters used.
\subsubsection{Cube-stacking}
We consider a cube-stacking environment consisting of $8$ cubes of different sizes and colors. The cubes, along with their attributes are listed as shown in Table \ref{Table:cubes}.
\begin{table}[h]
\centering
\begin{tabular}{||c c c||} 
 \hline
 Cube no. & Edge length & Color\\ [0.5ex] 
 \hline\hline
 1 & 5cm & Red\\ 
 \hline
 2 & 4cm & Red\\
 \hline
 3 & 3cm & Red\\
 \hline
 4 & 2cm & Red\\
 \hline
 5 & 10cm & Blue \\ 
 \hline
 6 & 8cm & Blue \\ 
 \hline
 7 & 6cm & Blue \\ 
 \hline
 8 & 2cm & Blue \\ 
 \hline
\end{tabular}
\caption{Cube edge lengths and colors.}
\label{Table:cubes}
\end{table}
We assume that the desired configuration of the stack is in the decreasing order of edge lengths, such that the correct order of cube numbers would be: $[5,6,7,1,2,3,4,8]$ or $[5,6,7,1,2,3,8,4]$. We assume each possible configuration (including partially complete configurations) as a state, represented by an array of cube numbers in sequence, as mentioned above. The environment is initialized with a null stack (no cubes in the stack) and at each time step, the agent chooses an action $a$ to pick from one of the available cubes to add to the top of the stack or to remove the top most element of the stack (if it exists). Each configuration $s$ is evaluated by an evaluation function $E$, which accounts for how many elements of the stack are in their correct position. That is, $E(s)=l_{s}C(s)$, where $l_s$ is the length of the current stack and $C(s)$ is the number of elements of the stack in the correct position. Hence, longer stacks with more elements in the correct position receive higher scores.
The reward $r(s,a)$ of the \emph{primary} RL agent is computed based on how much the new state $s'$ improved or worsened the previous state $s$:
\begin{equation}
    r(s,a)=E(s')-E(s)+b
\end{equation}
\noindent where $b$ is a bonus reward ($=1$) for correctly completing the full stack, otherwise set to $0$.

The \emph{secondary} RL agent assumes the same states as the \emph{primary} agent, with binary action of $0/1$ corresponding to not querying/querying the LLM. The reward for querying the LLM and obtaining the correct response is set as $1$, and for an incorrect response, it is $-1$. The reward for not querying the LLM is set as $0$. We consider $s_{LLM}$ to be the correct solution if $E(s_{LLM})>\delta$. We also obtain the LLM policy $\pi_{LLM}$ directly from $s_{LLM}$. E.g., if $s_{LLM}=[5,6,7,1,2,3,4,8]$ and the current state is $[5,6,7]$, the next action as per $\pi_{LLM}$ would be the one that selects cube $1$. 

\subsubsection{Image Completion}
The image completion environment consists of a target $10\times10$ binary image $I_T$ of a common, but arbitrary shape. The environment is initialized with an image $I$ with all pixels set to $0$, and a \emph{primary} DQN \cite{mnih2015human} agent is tasked with learning a policy to determine the configuration of individual pixels in order to produce $I_T$. The state consists of the vectorized representation of the image concatenated with the horizontal and vertical positions of the pixel under consideration. The agent has a binary action space, where it can choose to either change the pixel value (to $0$ if the current pixel value is $1$ and to $1$ if the current pixel value is $0$) or retain it as it is. In each episode of learning, each pixel in the image is subjected to the RL agent's actions until the episode terminates. An image's closeness to the target image is measured as:

\begin{equation}
E(I)=\sum_{i=0}^{N}\frac{1.(p_{i}={p_{T}}_{i})}{N}
\end{equation}

\noindent where $1.$ is an indicator function, $p_i$ is the intensity of the $i^{th}$ pixel in the constructed image and ${p_{T}}_{i}$ is the corresponding pixel intensity in the target image, and $N$ is the total number of pixels. At each step, the agent's reward is determined as:

\begin{equation}
r(s,a)=E(I')-E(I)+b   
\end{equation}
\noindent where action $a$ changes $I$ to $I'$, and a bonus $b$ is obtained if $E(I')>\delta$, indicating $I'$ is a desirable solution. By comparing the individual pixels in the LLM solution $I_{LLM}$ with the partially learnt image $I$, we determine the actions ($\pi_{LLM}$: change pixel intensity or retain it) for each pixel in $I$.


The \emph{secondary} agent takes $I$ as the state, has a binary action space, and a reward function that produces $1$ for correct LLM  extrapolations of the target image, and $0$ otherwise.

\begin{figure}[h]
\centering
    \includegraphics[width=0.7\columnwidth]{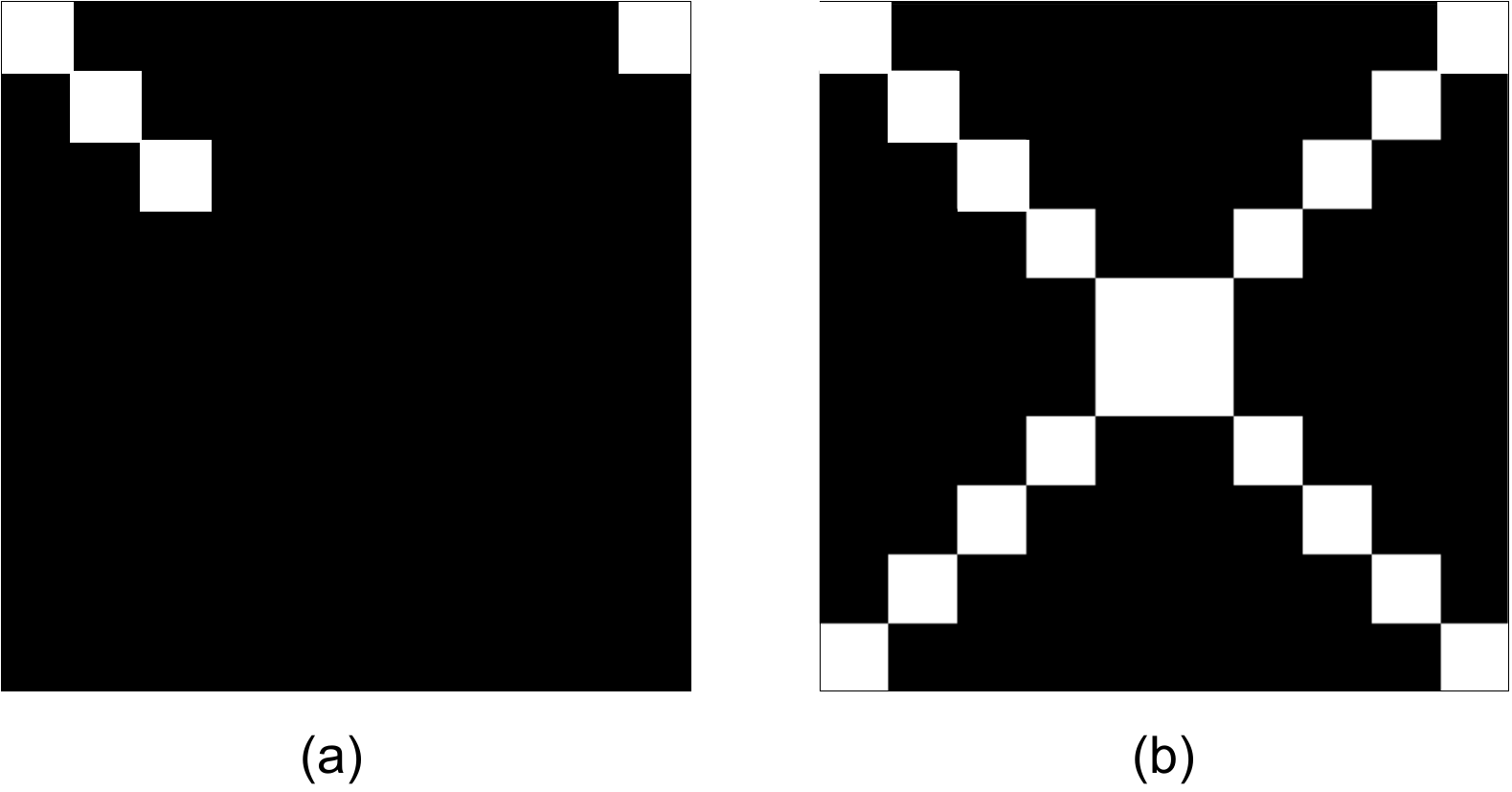}
    \caption{(a) An example of a partially learned image and (b) the target image in the Image Completion task.}
    \label{sample_imagecompletion}
\end{figure}

\subsubsection{Object Arrangement}

\begin{figure}[h]
\centering
    \includegraphics[width=0.5\columnwidth]{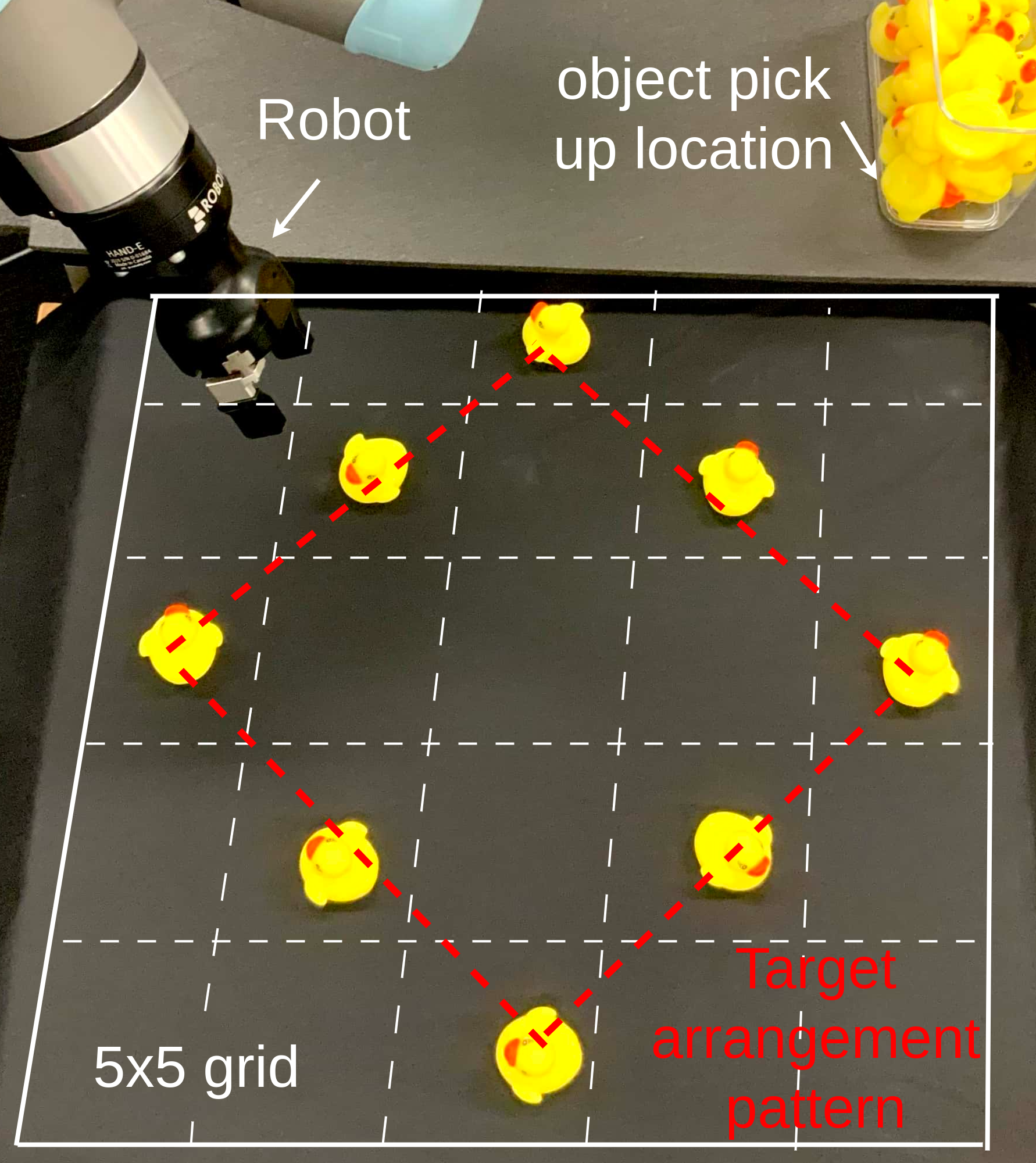}
    \caption{Experiment setup for the object arrangement task, which involves a $5\times5$ grid with the target arrangement pattern (diamond pattern) shown in red.}
    \label{arrangement_task}
\end{figure}

In the object arrangement task, we use a UR5e\footnote{https://www.universal-robots.com/products/ur5-robot/} robot arm with an end effector to arrange objects in a common arrangement patterns (a diamond pattern is used in the experiments) on a $5\times5$ grid as shown in Figure \ref{arrangement_task}. The grid is represented as a $5\times5$ matrix with `1's and `0's to respectively indicate the presence and absence of objects in a grid position. The environment is initialized with no objects on the grid, and begins with a DQN agent attempting to pick objects one at a time from the object pick up location, and place them on the grid to form the desired diamond pattern (Figure \ref{arrangement_task}). The \emph{primary} agent's state consists of a concatenated vector of the current arrangement matrix in vectorized form, and the horizontal and vertical coordinates of a grid position under consideration. The \emph{primary} agent's action space is binary, corresponding to actions to either drop or not drop the object on to the grid. The arrangement configuration after action execution is used to construct the next state. The \emph{primary} reward function is such that each action that improves the configuration (makes it closer to the target arrangement as per an evaluation function) is associated with a small positive reward, with other actions associated with $0$ reward. As the arrangement is represented as a matrix, we use the same evaluation and reward function as used in the Image Completion task. The agent receives a reward after an action is decided, but object dropping actions are executed only when positive rewards are obtained. A bonus reward is obtained when the arrangement pattern exactly matches the target pattern, after which the episode terminates. An episode may also terminate if the target pattern is not found within the specified episode horizon. After each episode, the objects are removed from the table and learning continues.     

As in the other tasks, the \emph{secondary} agent is fed the partial patterns encountered during learning (in the form of a matrix) and tasked with deciding whether to query the LLM, which suggests solutions in the form of $5\times5$ matrix configurations corresponding to possible target patterns.

\subsection{Can LLMs Reliably Extrapolate Patterns?}
\begin{figure}[h]
\centering
    \includegraphics[width=0.8\columnwidth]{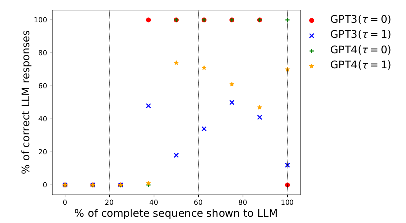}
    \caption{Average accuracy (over $100$ queries for each data point) of the LLM (GPT3 and GPT4) predictions for different temperatures ($\tau=0$ and $\tau=1$) when different fractions of the complete sequence are used to prompt the LLM.}
    \label{LLMaccuracy}
\end{figure}

We first characterize the accuracy of LLM predictions in the cube stacking task by prompting it with partially complete states and measuring the accuracy of the obtained LLM predictions for different temperatures. The temperature $\tau$ ($0\leq\tau\leq1$) of the LLM is a controllable parameter that determines the randomness of the LLM response (lower the $\tau$, the more deterministic is the LLM response). 
 As seen in Figure \ref{LLMaccuracy}, for low $\tau$ ($\tau=0$), the LLMs (both GPT3 and GPT4) are unable to predict the correct sequence from very short partial sequences. However, once $>35-45\%$ of the sequence is included as part of the prompt, the LLMs are able to correctly predict the full sequence. An interesting exception to this is that when the full sequence is shown to the LLM. In this case, GPT3 ($\tau=0$) is unable to reproduce the sequence ($0\%$ accuracy), while GPT4 ($\tau=1$) is more robust ($100\%$ accuracy). This perhaps highlights the limitations of the predictive capacity of GPT3 relative to GPT4. For higher LLM temperatures ($\tau=1$), we note that similar to the case with low temperature, the LLMs (both GPT3 and GPT4) are unable to make correct predictions when prompted with short partial sequences. However, when longer sequences are used for prompting, the prediction accuracy improves, but is not uniform with respect to the sequence length. 

These results firstly indicate that LLMs can indeed extrapolate the target pattern using partial sequences in the prompt, confirming our basic intuitions regarding LLMs' extrapolation abilities, which forms the basis for LaGR. Secondly, the non-uniform nature of these plots reaffirm our intuitions that there exist certain partial patterns from which it may be more beneficial to query the LLM. This is precisely what SEQ aims to learn, following which it ensures that unnecessary queries are avoided, and that the agent is only queried in states where the LLM is more likely to produce the target pattern. With these intuitions confirmed, we proceed to the empirical evaluations of LaGR-SEQ.

\subsection{Sample-efficiency via LaGR}
\begin{figure*}[h]
\centering
    \includegraphics[width=0.79\textwidth]{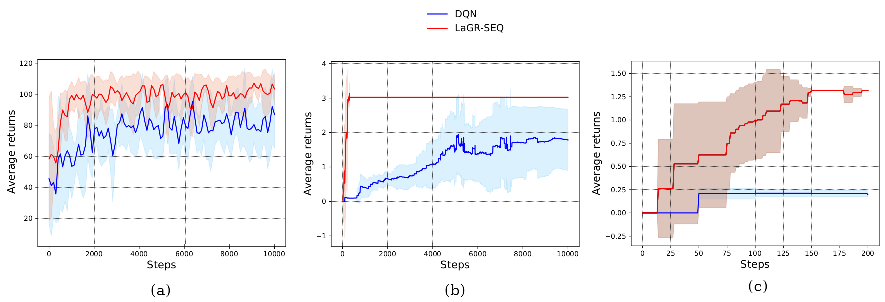}
    \caption{Average performances of DQN vs LaGR-SEQ in the (a) Cube stacking, (b) Image Completion and (c) Object Arrangement environments across $10, 10$ and $5$ trials respectively.}
    \label{llmperformanceall}
\end{figure*}
We deploy the LaGR-SEQ agent (which uses $Q-$learning along with GPT3 for cube stacking and DQN with GPT4 for the other environments) in the environments described and compare its performance with that of a standard RL agent. For these experiments, we use $P=1$, deterministically following $\pi_{LLM}$ once the correct solution is found. As seen in Figure \ref{llmperformanceall}, our LaGR-SEQ agents learn with considerably fewer interactions in all three tasks, demonstrating the improved sample efficiency offered by LaGR. We note that the improved sample efficiency observed occurs due to LaGR alone (both LaGR and LaGR-SEQ perform equivalently - relevant experiments are included in the Appendix), and SEQ only serves to improve querying efficiency, as it is designed to do so. Due to the trial and error nature of RL, there is a chance that identical queries are repeatedly made to the LLM. In order to avoid the resulting unnecessary costs, we use a caching mechanism to handle this, the details of which we include in the Appendix. 

As we performed the object arrangement task on a real robot, we limited our experiments to $5$ trials, each of $200$ steps (Figure \ref{llmperformanceall} (c)). Even so, we observe that LaGR-SEQ achieves a high reward with just over $100$ real world interactions\footnote{See video \href{https://www.dropbox.com/scl/fi/1xvge1dzjsip8hoirm6ju/LLM-guided-final_vid.mp4?rlkey=1mrx0vx8nfll1ajj4k2elidek&dl=0}{here}}. This demonstrates the suitability of LaGR-SEQ agents to be deployed in real world tasks directly, without a simulator. For completeness, we conducted extended versions of this experiment in simulation (see Appendix), which show that the DQN performance improves, albeit with more environment interactions.

\begin{figure*}[h]
\centering
    \includegraphics[width=0.8\textwidth]{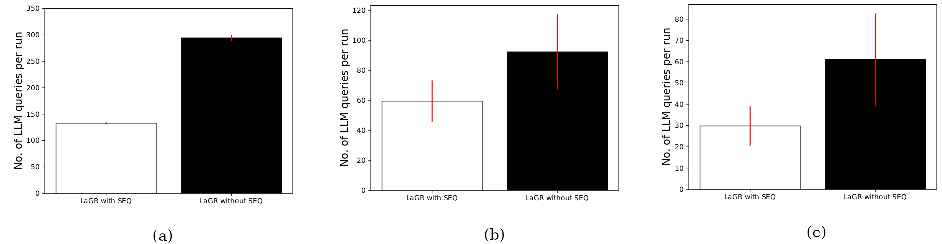}
    \caption{Number of LLM queries with and without SEQ in (a) Cube stacking, (b) Image Completion and (c) Object Arrangement environments across $10,10$ and $5$ trials respectively. Error bars indicate standard error.}
    \label{llmperformance-SEQ}
\end{figure*}

\subsection{Reducing LLM Queries via SEQ}
In addition to the improved sample efficiency, we show in Figure \ref{llmperformance-SEQ} the reduction in the number of LLM queries per trial with and without SEQ. We observe that agents with SEQ query the LLM substantially less frequently compared to those without SEQ in all the environments described.

\subsubsection{SEQ and LLM temperature $\tau$:} In further experiments in the cube stacking environment, we observe that as $\tau$ increases, the agents make more LLM queries, as shown in Table \ref{Table:LLMtemps}, although SEQ helps reduce the total queries.

\begin{table}[h]
\centering
\begin{tabular}{||c c c||} 
 \hline
 $\tau$ & LaGR-SEQ & LaGR (without SEQ)\\ [0.5ex] 
 \hline\hline
 0 & 126$\pm$2.94 & 284$\pm$0.94\\ 
 \hline
 0.5 & 133$\pm$1.41 & 294.3$\pm$3.2\\
 \hline
 1 & 138.33$\pm$7.31 & 331.66$\pm$1.44\\
 \hline
\end{tabular}
\caption{Mean$\pm$std error of no. of LLM queries per run (over $10$ runs) with and without SEQ in the cube stacking environment for different LLM temperatures $\tau$.}
\label{Table:LLMtemps}
\end{table}

\subsubsection{SEQ and LLM type:} The effectiveness of SEQ is also significantly affected by the type of LLM used. In Table \ref{Table:LLMtypes} corresponding to the image completion task, we observe that although  SEQ leads to a considerable reduction in the number of queries for both GPT3 and GPT4, the absolute number of samples used is much higher for GPT3. This is indicative of GPT4's superior performance over GPT3 for this task. 

\begin{table}[h]
\centering
\begin{tabular}{||c c c||} 
 \hline
 LLM & LaGR-SEQ & LaGR (without SEQ)\\ [0.5ex] 
 \hline\hline
 GPT3 & 236.4$\pm$33.21 & 389.9$\pm$78.62\\ 
 \hline
 GPT4 & 59.5$\pm$13.79 & 92.3$\pm$25.03\\
 \hline
 \end{tabular}
\caption{Mean$\pm$std error of no. of LLM queries per run (over $10$ runs) with and without SEQ in the Image Completion task for GPT3 and GPT4.}
\label{Table:LLMtypes}
\end{table}

\section{Discussion and Limitations}

We showed that LaGR-SEQ is a powerful approach to combine the context aware LLMs with RL agents for pattern completion tasks. However, we note that its applicability is restricted to tasks where the target pattern is relatively common and semantically meaningful, without which LLMs would fail to extrapolate the desired solutions. In such a case, the sample efficiency of LaGR would simply drop to that of the underlying RL agent. Also, since the framework achieves sample-efficiency by relying on LLM's extrapolated output, LaGR is suited to long sequence problems with easily extrapolatable patterns. Preliminary experiments (refer to Appendix) support this intuition. 
 We also note that our approach is limited to environments where the LLM policy can be easily extracted from the extrapolated solution. For some environments, this aspect may not be straightforward. Next, in our experiments, LLM queries are triggered following a positive reward. While convenient and logical, this type of query scheduling is inherently dependent on the reward design, which could lead to unanticipated/undesired querying characteristics (e.g. too frequent/too rare). Hence, we believe there is value in investigating more robust scheduling strategies. Lastly, in this work, we have only considered GPT3 and GPT4 due to their relatively superior performance for our tasks. However, we acknowledge that extending this work to other LLMs, particularly open-source variants is important, and in the interest of open and free research.

\section{Conclusion}
We introduced LaGR-SEQ, a novel framework for improving the sample efficiency of RL with context-aware guidance via LLMs. Our framework also simultaneously learns when it is appropriate to make LLM queries, thereby preventing unnecessary querying of LLMs, which can be slow and expensive. We described the components of our approach in detail and demonstrated its benefits in various pattern completion tasks, where it was shown to improve learning performance while exhibiting improved query-efficiency. We believe our proposed framework would benefit future works aiming to leverage contextual information from LLMs for efficient RL training in pattern completion tasks. 
\newpage     
\bibliographystyle{plainnat}
\bibliography{aaai24}

\section{Task Descriptors and Prompts}
We present below the task descriptors used for the tasks described in this work. Each task descriptor is  combined with partial patterns (which are inserted as indicated) to form the prompt.

\subsubsection{Cube Stacking:}
\begin{verbatim}
A table contains the following objects: \
- Red cube of edge length 5cm \
(represented by 'a')\
- Red cube of edge length 4cm \
(represented by 'b')\
- Red cube of edge length 3cm\
(represented by 'c')\
- Red cube of edge length 2cm\
(represented by 'd')\
- Blue cube of edge length 10cm\ 
(represented by 'e')\
- Blue cube of edge length 8cm\
(represented by 'f')\
- Blue cube of edge length 6cm\
(represented by 'g')\
- Blue cube of edge length 2cm\ 
(represented by 'h')\
A human is currently stacking some\
of the cubes in the following \
sequence (from bottom to top): \
\end{verbatim}
\textit{Insert Partial Pattern here}
\begin{verbatim}
You are an organizing robot.\
Stack the remaining cubes in the\
pattern that human seems to be \
following. The final stack should\
have all 8 cubes. Lets think step by step.\
For example, if the stack is initially\
in the order (bottom to top)\
['a','d','b']\
Then, the explanation is that\ 
the human is likely first stacking the\
red cubes irrespective of their 
edge lengths, followed by the \
blue cubes. So a possible order \
would be: (bottom to top):\
['a','d','b','c','f','e','g','h']\
Make sure your response contains \
only the order in the form of a \
list and not the explanation.
\end{verbatim}

\subsubsection{Image Completion:}

\begin{verbatim}
The following matrix corresponds to an \
image of size 10x10. 1 represents a white\
pixel, while 0 represents a black pixel.\
If the initial image is:\
\end{verbatim}
\textit{Insert Partial Pattern here}
\begin{verbatim}
Observe the pattern in the initial image.\
It is a partially complete pattern of a \
common geometric shape. Guess the final \
shape and display it in the form of a \
10x10 matrix. Note that all the initial\
pixels are in their correct positions. \
So the final image must necessarily \
contain the initial pixels as they are,\
and should not contain any additional\
pixels that are not part of the shape.\
Lets think step by step. Make sure \
your response contains only the\
image, and not the explanation.\
Example 1: If the initial image is: \
    [[0,0,0,0,0,0,0,0,0,0],
    [0,0,0,0,0,0,0,0,0,0],
    [0,0,0,1,1,1,0,0,0,0],
    [0,0,1,0,0,0,0,0,0,0],
    [0,0,1,0,0,0,0,0,0,0],
    [0,0,1,0,0,0,0,0,0,0],
    [0,0,1,0,0,0,0,0,0,0],
    [0,0,0,1,0,0,0,0,0,0],
    [0,0,0,0,0,0,0,0,0,0],
    [0,0,0,0,0,0,0,0,0,0]]\
    Then the full image may be\
    in the shape of an oval, in which\
    case the final image would be:\
    [[0,0,0,0,0,0,0,0,0,0],
    [0,0,0,0,0,0,0,0,0,0],
    [0,0,0,1,1,1,0,0,0,0],
    [0,0,1,0,0,0,1,0,0,0],
    [0,0,1,0,0,0,1,0,0,0],
    [0,0,1,0,0,0,1,0,0,0],
    [0,0,1,0,0,0,1,0,0,0],
    [0,0,0,1,1,1,0,0,0,0],
    [0,0,0,0,0,0,0,0,0,0],
    [0,0,0,0,0,0,0,0,0,0]]\
    Example 2: If the initial \
    configuration is: \
    [[0,0,0,0,0,0,0,0,0,0],
    [0,0,0,0,0,0,0,0,0,0],
    [0,0,0,0,1,0,0,0,0,0],
    [0,0,0,1,0,1,0,0,0,0],
    [0,0,1,0,0,0,1,0,0,0],
    [0,0,0,0,0,0,0,0,0,0],
    [0,0,0,0,0,0,0,0,0,0],
    [0,0,0,0,0,0,0,0,0,0],
    [0,0,0,0,0,0,0,0,0,0],
    [0,0,0,0,0,0,0,0,0,0]]\
    Then a possible pattern \
    could correspond \
    to a triangle, such that the \
    final pattern would be:\
    [[0,0,0,0,0,0,0,0,0,0],
    [0,0,0,0,0,0,0,0,0,0],
    [0,0,0,0,1,0,0,0,0,0],
    [0,0,0,1,0,1,0,0,0,0],
    [0,0,1,0,0,0,1,0,0,0],
    [0,1,0,0,0,0,0,1,0,0],
    [1,1,1,1,1,1,1,1,1,0],
    [0,0,0,0,0,0,0,0,0,0],
    [0,0,0,0,0,0,0,0,0,0],
    [0,0,0,0,0,0,0,0,0,0]]
\end{verbatim}
\subsubsection{Object Arrangement:}
\begin{verbatim}
The following matrix corresponds to \
positions on a 5x5 table. 1 \
represents the presence of an \
object, while 0 represents an \
empty space on the table.\
If the initial table \
configuration is:
\end{verbatim}
\textit{Insert Partial Pattern here}
\begin{verbatim}
Observe the pattern of objects \
in the initial configuration.\
It is a partially complete \
pattern of a common scheme \
in which objects would be arranged.\
Guess the final arrangement scheme\
and display it in the form of\
a 5x5 matrix. Note that all the\
initial object positions are correct.\
So the final configuration must\
necessarily contain the initial\
object positions as they are.\
Lets think step by step. Make \
sure your response contains \
only the configuration, and\
not the explanation.\
Example 1: If the initial \
arrangement configuration is:\
    \textit{[[0,0,0,0,0],
    [0,1,0,1,0],
    [0,0,0,0,0],
    [0,1,1,0,0],
    [0,0,0,0,0]]}\\
Then the full arrangement may\
be in the shape of a square, \
in which case the final \
arrangement would be:}\
    [[0,0,0,0,0],
    [0,1,1,1,0],
    [0,1,0,1,0],
    [0,1,1,1,0],
    [0,0,0,0,0]]\
Example 2: If the initial \
arrangement is: \
    [[0,0,1,0,0],
    [0,1,0,0,0],
    [0,0,0,1,0],
    [0,0,1,0,0],
    [0,0,0,0,0]]\
Then a possible arrangement\
could be in the shape of an \
oval, such that the final \
arrangement would be:\
    [[0,0,1,0,0],
    [0,1,0,1,0],
    [0,1,0,1,0],
    [0,0,1,0,0],
    [0,0,0,0,0]]\
\end{verbatim}

\section{Effect of LLM Temperature on LaGR}
\begin{figure}[h]
    \includegraphics[width=\columnwidth]{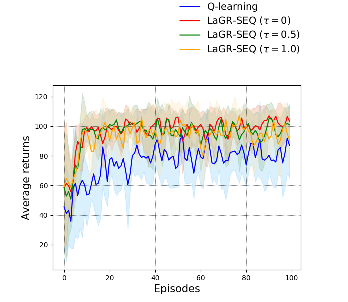}
    \caption{Comparison of average returns (over $10$ trials) vs episodes of our approach for different LLM temperatures ($\tau=0$, $\tau=0.5$ and $\tau=1.0$) versus $Q-$learning in the cube stacking environment. Shaded regions represent one standard deviation.}
    \label{llmperformance_lagrtemp}
\end{figure}

Although we discussed in the main text that the LLM temperature had a considerable influence on SEQ, as observed in Figure \ref{llmperformance_lagrtemp}, LaGR does not seem to be greatly affected by the LLM temperature $\tau$ in our experiments with cube stacking. Similar trends were also observed in other environments.

\section{Effect of LLM type on LaGR}
\begin{figure}[h]
    \includegraphics[width=\columnwidth]{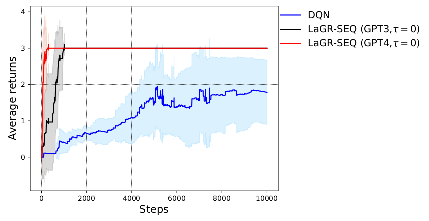}
    \caption{Comparison of average return (over $10$ trials) of LaGR-SEQ using GPT4,($\tau=0$) and GPT3,($\tau=0$) vs DQN for the Image Completion environment. Shaded regions represent one standard deviation.}
    \label{llmperformance_lagrtype}
\end{figure}

The choice of LLM appeared to slightly influence LaGR's performance as seen Figure \ref{llmperformance_lagrtype} in the image completion task. This is primarily due to the superior responses from GPT4 compared to GPT3. However, as GPT was also able to extrapolate the required pattern, albeit with more queries (as shown in Table \ref{Table:LLMtypes} in the main text), the \emph{primary} agents performance did not suffer drastically.

\section{Performance of LaGR with Stack Size}

We repeated the cube-stacking experiments on stacks of different sizes, the stack size being determined by the total number of cubes under consideration. We found that with larger stack sizes, the ratio of the performance of LaGR to that of $Q-$ learning, which we term as \emph{performance ratio}, shows an increasing trend, as seen in Table \ref{Table:perfratio}. This is due to the fact that the LLM in LaGR essentially extrapolates the partially completed pattern. Hence, as long as the extrapolation is based on an accurate context, LaGR's solution would remain valid irrespective of the stack size. On the other hand, the performance of $Q-$learning, which learns devoid of any context, depends on the stack size. The increasing performance ratio with stack size confirms our intuitions, initially discussed in the section `Discussion and Limitations' regarding the the nature of LaGR to perform better on longer sequences and easily exptrapolatable patterns.
\begin{table}[h]
\centering
\begin{tabular}{||c c||} 
 \hline
Stack size & Performance ratio\\ [0.5ex] 
 \hline\hline
5 & $1.05\pm0.01$\\ 
 \hline
8 & $1.28\pm0.04$\\
 \hline
11 & $1.33\pm0.19$\\
 \hline

 \hline
\end{tabular}
\caption{Average $\pm$ standard error of performance ratios (ratio of total returns accumulated by LaGR to that accumulated by $Q-$learning) with stack size (total number of cubes in the stack) for the cube stacking environment.}
\label{Table:perfratio}
\end{table}

\section{LaGR-SEQ vs LaGR}
\begin{figure}[h]
    \includegraphics[width=\columnwidth]{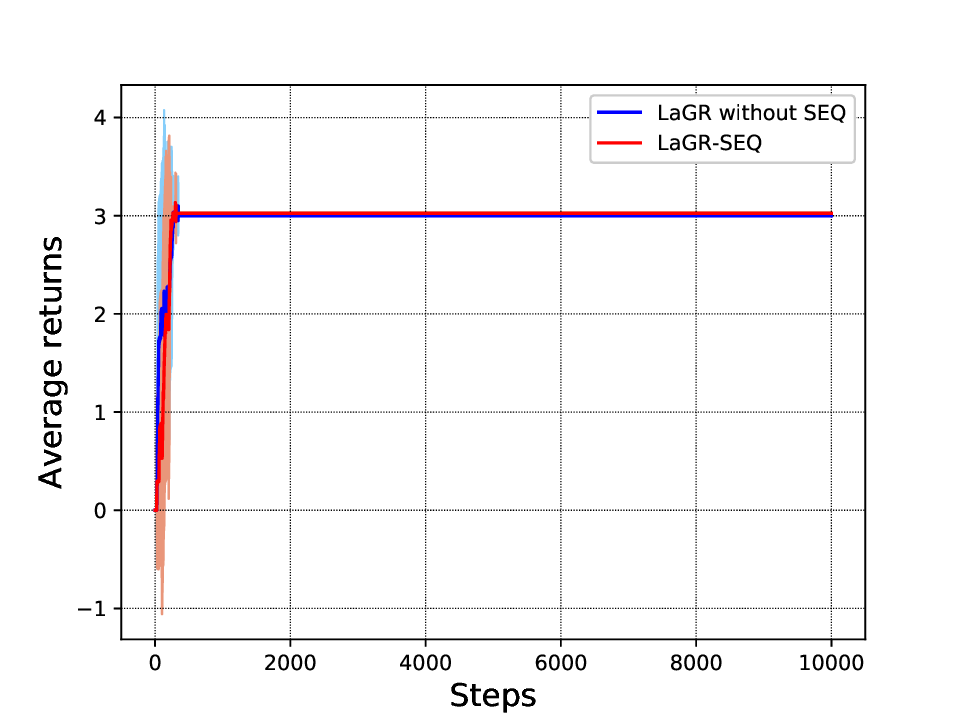}
    \caption{Comparison of average returns (over $10$ trials) of LaGR with and without SEQ in the Image Completion task. Shaded regions represent one standard deviation. As seen from the figure, SEQ does not significantly influence the learning efficiency.}
    \label{seqvnoseq}
\end{figure}

We demonstrated through various experiments the effectiveness of LaGR for sample-efficient learning. Although SEQ was proposed as a mechanism to reduce the number of LLM queries, we contend that it only influences querying efficiency (as shown in the main paper), and has no influence on the primary agent's sample efficiency. We demonstrate this in Figure \ref{seqvnoseq}, which shows that a well learnt SEQ policy can reduce LLM querying with no noticeable reduction in performance.

\section{Simulated Object Arrangement}
\begin{figure}[h]
    \includegraphics[width=\columnwidth]{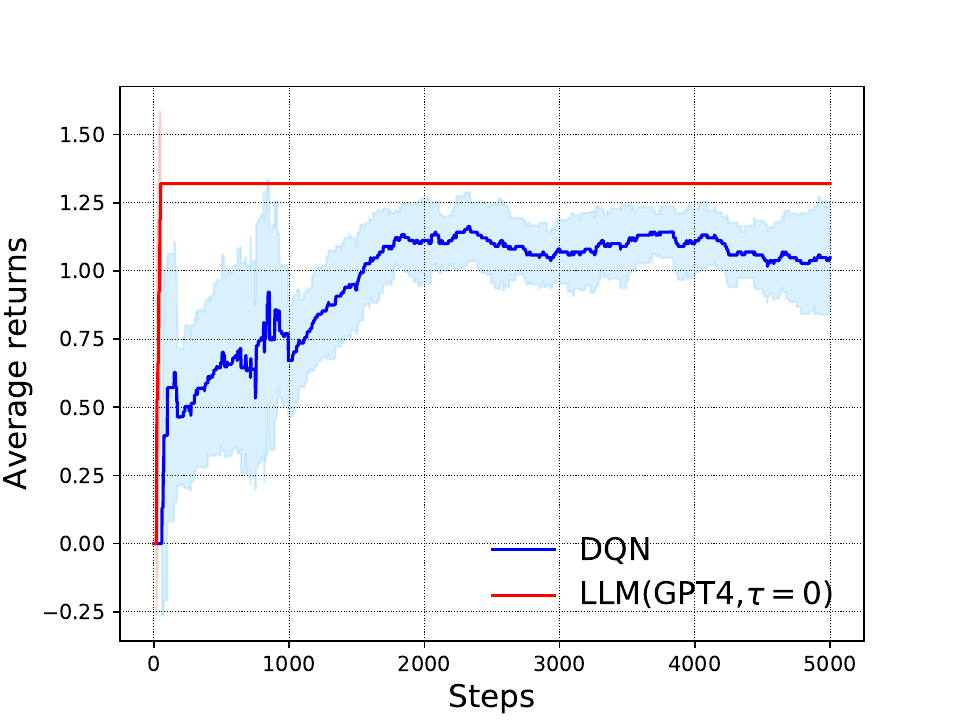}
    \caption{Comparison (over $10$ trials) of the performance of LaGR-SEQ using GPT4,($\tau=0$)  vs DQN for an extended simulated version of the object arrangement task.}
    \label{extendedrobot}
\end{figure}

Figure \ref{extendedrobot} shows an extended version of the object arrangement task carried out in simulation. As observed from the figure, DQN also achieves a good performance towards the end of training.
 
\section{Non-binary Secondary Agent Rewards}
\begin{figure}[h]
    \includegraphics[width=\columnwidth]{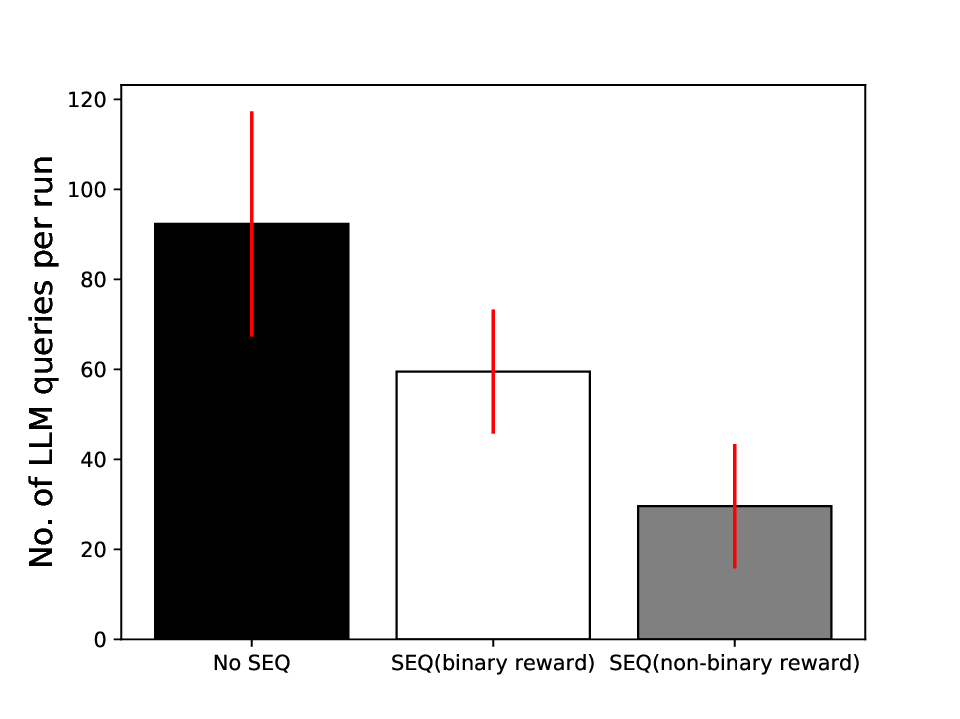}
    \caption{Comparison (over $10$ trials) of the number of LLM queries per run without SEQ, with SEQ (binary reward) and with SEQ (non-binary reward).}
    \label{binvnobin}
\end{figure}

Although in our experiments, we provided binary rewards to our \emph{secondary} RL agent (SEQ), our framework as such does not restrict these rewards to be binary. In fact, using domain knowledge, we found that non-binary rewards were able to achieve a much superior querying performance, as shown in Figure \ref{binvnobin}. The reward was specified as a logistic function, designed based on our domain knowledge of what constituted a close match to the target pattern:
\begin{equation*}
 r=\frac{1}{(1+e^{(-p_{1}{((E(s_{LLM})/N)-p2)))}}}   
\end{equation*}
\noindent where $p_1=20$, $p_2=0.9$, $N=100$ (total number of pixels in image).

\section{Implementation Details}
In this section, we outline the various implementation details for LaGR-SEQ. Apart from the hyperparameters summarized in the next subsection, we note that our experiments were conducted on CPUs 2x AMD EPYC 7742 (64 cores @ 2.25GHz) and 4TB RAM and on average, consumed a wall-clock time of about 30-60 min per trial depending on the state of caching (explained later) and the environment. The robotics experiment was conducted on a system: AMD 1900x8-core processor x 16 with 32 GB RAM and an NVIDIA GeForce RTX2080Ti graphics card. The wall-clock time for the robotics experiments was about 22-24 minutes per trial. We now list the various hyperparameters used in our experiments.
\subsection{Hyperparameters}
We present in Table \ref{Table:hyperparams} the various hyperparameters used in our experiments.

\begin{table*}
\centering
\begin{tabular}{||c c c c||} 
 \hline
 Hyperparameter & Cube Stacking & Image Completion& Object Arrangement\\ [0.5ex] 
 \hline\hline
  learning rate $\alpha$ (primary)& $0.1$& $1e-3$&$1e-3$\\ 
 \hline
 learning rate $\alpha$ (secondary)& $0.1$& $1e-3$&$1e-3$\\ 
 \hline
  discount factor $\gamma$ (primary)& $0.95$  & $0.95$&$0.95$\\ 
 \hline
   discount factor $\gamma$ (secondary)& $0.95$ & $0.95$&$0.95$\\ 
 \hline
    Reward bonus& $1$ & $1$&$1$\\ 
 \hline
   Episode horizon $H$& $100$  & $500$ &$50$\\ 
 \hline
   Evaluation threshold $\delta$ & $1$ & $0.95$&$0.99$\\ 
\hline
   Initial $\epsilon$ & $1$ & $1$&$1$\\ 
 \hline
 Minimum $\epsilon$ & $0.05$ & $0.1$&$0.1$\\ 
 \hline
 $\epsilon$ decay & Linear & exponential-decay factor:$0.998$&exponential-decay factor:$0.998$\\ 
 \hline
   NN achitecture& n/a & 2 hidden layers (128 neurons each)&2 hidden layers (64 neurons each)\\
   &  & linear output layer&linear output layer\\
 \hline
    Hidden layer activation& n/a & relu&relu\\ 
    \hline
   Optimizer& n/a & Adam&Adam\\ 
   \hline
    Loss& n/a & mse&mse\\ 
   \hline
   Batch size& n/a & 32&16\\
     \hline
   Replay buffer size& n/a & $10000$&$10000$\\
 \hline
  \hline
\end{tabular}
\caption{Summary of hyperparameters used in our experiments.}
\label{Table:hyperparams}
\end{table*}
\subsection{Caching LLM Responses}

As we make LLM queries in an RL task, we contend that it is possible for identical queries to be made repeatedly. This could greatly increase the expenses of conducting experiments, as well as slow down the speed of experiments (owing to time it takes for the LLM to respond from the server). In order to mitigate these issues, we cached partial states along with their associated LLM query responses in a cache file which was called at the beginning of each experiment run. For each query state that appeared in the experiment, the existence of that state in the cache was first checked, and if the state happened to already exist in the cache, its corresponding cached LLM response was used instead of actually querying LLM from the server. LLM queries were made to the server only for query states that did not exist in the cache. After such a query is made, the query state, along with the LLM response was added to the cache, and the process continued until training was completed. For experiments with LLM temperatures $\tau>0$, the LLM response is not deterministic. In order to account for this, for each query state, we performed $10$ queries to the server, and all of these responses were stored in the cache file. If this state appeared again, then it would be found in the cache, and one of the cached LLM responses was picked as the corresponding LLM response. Although this method of caching required several queries to be made to the server at the beginning of an experiment, once a majority of the responses are cached, very few responses would need to be made for subsequent experiments. We also note that this approached required us to build separate caches for each experiment and for each LLM temperature setting. 

\subsection{Robot Experiment Details}

For the \emph{Object Arrangement} task a \emph{UR5e} robot arm \footnote{https://www.universal-robots.com/products/ur5-robot/} was used as the manipulator. To operate the arm from a workstation, \emph{Real-Time Data Exchange (RTDE)} protocol \footnote{https://www.universal-robots.com/articles/ur/interface-communication/real-time-data-exchange-rtde-guide/}, implemented by \emph{ur\_rtde} client library \footnote{https://sdurobotics.gitlab.io/ur\_rtde/} was used.

The task was set up as placing objects on a table surface with a grid of $5\times5$ cells, each of which was a square of size $5$ cm.  The input for the task then was taken as grid coordinates of a cell, e.g., ($2$, $3$) for the cell of the second row and the third column. Then these grid coordinates were translated to the cell positions in terms of metric unit (metres) distances from the origin ($0$, $0$) cell on the table surface. Thereafter, these 2-dimensional coordinates were transformed to 3-dimensional coordinates in the robot base frame using \emph{ROS TF2} library \footnote{http://wiki.ros.org/tf2}. In order to achieve this transformation, a static transformer (static\_transform\_publisher) was set up specifying the distance and rotation to the table surface from the robot base. To facilitate communication with this program running on the \emph{UR5e} workstation from the LaGR-SEQ program, a \emph{gRPC} based server and client stub was generated \footnote{https://grpc.io}.

\section{Expenditure for Querying GPT}
The LLM queries in this work were performed via the OpenAI API with a paid account. The cumulative expenditures for all experiments, including during development as well as testing of our approach totalled to US\$109.42. We note that this amount would be substantially larger without the caching system we described earlier.
\end{document}